**Brief Title**

Physics Symbolic Learner for Ground Motion Model and Modeling

**Author information**

1.  Su Chen, Key Laboratory of Urban Security and Disaster Engineering of the Ministry of Education, Beijing University of Technology, Beijing, 100124, China
2.  Xianwei Liu, Institute of Geophysics, China Earthquake Administration, Beijing,100081, China
3.  Lei Fu (corresponding author, fulei11@cea-igp.ac.cn), Institute of Geophysics, China Earthquake Administration, Beijing,100081, China
4.  Suyang Wang, Key Laboratory of Urban Security and Disaster Engineering of the Ministry of Education, Beijing University of Technology, Beijing, 100124, China
5.  Bin Zhang, Institute of Geomechanics, Chinese Academy of Geological Sciences, Beijing, 100081, China
6.  Xiaojun Li, Key Laboratory of Urban Security and Disaster Engineering of the Ministry of Education, Beijing University of Technology, Beijing, 100124.

**Acknowledgments**

The authors thank the Pacific Earthquake Engineering Research Center (PEER) for sharing the Next Generation Attenuation-West2 Project (NGA-West2) flatfile. This work is under the support of National Natural Science Foundation of China (Grant Numbers 52192675 and 52278352), National Key R&D Program of China (Grant Number 2022YFC003503), Special Fund of Institute of Geophysics, China Earthquake Administration (Grant Number DQJB22B28).

**Data availability**

The database of NGA-West2 for the present study was retrieved from the Pacific Earthquake Engineering Research Center (https://peer.berkeley.edu/research/data-sciences/databases, last accessed May 2022).

**Conflict of interest**

The authors acknowledge that there are no conflicts of interest recorded.

# Physics Symbolic Learner for Discovering Ground-Motion Models Via NGA-West2 Database

**Abstract**

Ground-motion model (GMM) is the basis of many earthquake engineering studies. In this study, a novel physics-informed symbolic learner (PISL) method based on the Nest Generation Attenuation-West2 database is proposed to automatically discover mathematical equation operators as symbols. The sequential threshold ridge regression algorithm is utilized to distill a concise and interpretable explicit characterization of complex systems of ground motions. In addition to the basic variables retrieved from previous GMMs, the current PISL incorporates two a priori physical conditions, namely, distance and amplitude saturation. GMMs developed using the PISL, an empirical regression method (ERM), and an artificial neural network (ANN) are compared in terms of residuals and extrapolation based on obtained data of peak ground acceleration and velocity. The results show that the inter- and intra-event standard deviations of the three methods are similar. The functional form of the PISL is more concise than that of the ERM and ANN. The extrapolation capability of the PISL is more accurate than that of the ANN. The PISL-GMM used in this study provide a new paradigm of regression that considers both physical and data-driven machine learning and can be used to identify the implied physical relationships and prediction equations of ground motion variables in different regions.

## 1. Introduction

Ground-motion model (GMM) is among the most important models in seismic hazard assessment and have been used to predict the engineering-interested intensity measurements (IM) of ground motion for earthquake events. Based on seismological and geophysical observations as well as theoretical studies, a GMM is constructed using several empirical proxies to describe the intensity of the source spectrum, propagation attenuation, and site response. Sufficient ground-motion data are required to develop ground-motion prediction equations (GMPEs). These proxies include, but are not limited to, magnitude scaling, distance, style-of-faulting, anelastic attenuation, geometrical spreading, and time-averaged shear wave velocity of the uppermost 30 m ($V_{S30}$)[1,2].

Douglas[3] summarized that nearly one thousand GMMs have been published since 1964. More than 81% of these models were developed using empirical models derived from ground-motion data, whereas others were based on simulated ground motions[4], the hybrid stochastic–empirical method[5], and the referenced-empirical approach[6]. Among these models, the five GMMs developed in theNext Generation Attenuation-West2 (NGA-West2) project are the most widely acknowledged owing to their elaborate functional forms and robust global ground-motion database[7-11]. These GMMs are typically used to perform region-specific hazard assessment, seismic hazard assessment for earthquake-resistant design code, and property damage estimation[12]. Empirical-based physical models can improve the prediction accuracy of statistical methods using limited data owing to the full consideration of physical laws. However, such physical models often presuppose a functional form, which often limits the ability to effectively model and measure the complex and unknown behaviors of ground motion.

Exploiting the rapid growth of global ground-motion data and the development of machine learning (ML) method, various ML methods have been used to develop GMPEs for ground-motion IMs. Table 1 summarizes the GMMs constructed using ML models with different datasets. These GMPEs can fit the data well and fully exploit the intrinsic connections between parameters. However, the formulas for these GMPEs are not interpretable. Furthermore, the extrapolation ability of most GMPEs is questionable.

Table 1 Application of machine learning in GMPE

| Method | Parameters | Research Area | Year |
|---|---|---|---|
| Genetic programming[13] | $M_w$, $R_{hyp}$, $V_{S30}$ | Turkey | 2009 |
| Simulated annealing[14] | $M_w$, $R_{hyp}$, $V_{S30}$, Fault type | NGA-West1 | 2011 |
| Genetic programming[15] | $M_w$, $R_{JB}$ $V_{S30}$, Fault type | NGA-West1 | 2011 |
| Support vector regression[16] | $M_w$, $R_{Clstd}$, $V_{S30}$, Fault type | NGA-West1 | 2012 |
| Lagramge[17] | $M_w$, $R_{JB}$, $V_{S30}$, Fault type | NGA-West1 | 2013 |
| ANN[18] | $M_w$, $R_{JB}$, $V_{S30}$, Fault type, Focal depth | Europe | 2014 |
| Neuro-fuzzy inference[19] | $M_w$, $R_{Clstd}$, $V_{S30}$, Fault type | NGA-West1 | 2016 |
| M5 model tree[20] | $M_w$, $R_{Clstd}$, $V_{S30}$, Fault type | NGA project | 2016 |
| ANN[21] | $M_w$, $R_{rup}$, $V_{S30}$, Fault type | NGA-West2 | 2018 |
| Classification and Regression tree[22] | $M_w$, $R_{Clstd}$, $V_{S30}$, Fault type | NGA-West1 | 2018 |
| Prefix Gene Expression Programming[23] | $M_w$, $R_{epi}$, $V_{S30}$, Rank angle | Iran, Turkey, Armenia, Georgia | 2018 |
| Multilayer Perceptron (MLP)[24] | $M_w$, $R_{rup}$, $V_{S30}$, Fault type | NGA-West1 | 2019 |
| Deep neural networks (DNN)[25] | $M_w$, $R_{Clstd}$, $V_{S30}$, Rank angle | NGA-West2 | 2019 |
| ANN[26] | $M_w$, $R_{hypo}$, $V_{S30}$ | Oklahoma, Kansas, and Texas | 2019 |
| Refined Second-Order DNN[27] | $M_w$, $R_{JB}$, $V_{S30}$, $Z_1$, $Z_{TOR}$, structural period, fault type, region | NGA-West2 | 2021 |
| Bayesian neural network[28] | $M_w$, $R_{rup}$, $V_{S30}$, Fault type, region, hypocentral depth | NGA-West2 and NGA-Sub | 2023 |

Notes: $M_w$ is moment magnitude. $R_{hyp}$ (km) is hypocentral distance; $R_{Clstd}$ (km) is Closest distance from the recording site to the ruptured fault area. $R_{JB}$ (km) is shortest horizontal distance from the recording site to the vertical projection of the rupture on the surface. $R_{epi}$ is distance from the recording site to epicentre. $R_{rup}$ is closest distance from the recording site to the ruptured fault area. $V_{S30}$ (m/s) is average shear wave velocity over the top 30 m of site. $Z_1$ is basin depth. $Z_{TOR}$ is depth to the top of rupture.

In the last decade, symbolic regression and symbol-based deep learning have been used to identify intricate dynamic systems and governing equations[29]. Rudy[30] used the sequential threshold ridge regression (STRidge) algorithm to identify data-driven governing equations using all conceivable forms of variables contained in the function library (e.g., constant, primary, exponential, and partial derivatives). Whereas conventional data-driven methods (e.g., artificial neural networks, ANN) focus on reconstructing errors and fitting data, the STRidge algorithm provides explicit interpretability in terms of governing equations and the original set of variables[31]. Some scholars applied STRidge in the fields of fluid flow[32], structural systems[31], and stochastic processes[33]. In this study, the method is extended to the derivation of ground motion model to form symbolic-learning GMM. Notably, we fully considered the existing physics-informed GMM (herein referred to as PISL-GMM) when constructing the function library such that the PISL-GMM reflect a priori knowledge and physical constraints.

Herein, we propose a novel framework for constructing PISL-GMM based on the STRidge algorithm. The PISL-GMM incorporates the near-distance saturation of ground-motion amplitudes.

Approximately 16000 peak ground acceleration (PGA) and velocity (PGV) from the NGA-West2 database are used to discover the GMM and develop the GMPE. The predictive power of PISL-GMM is illustrated by comparing the residuals, standard deviation (SD) of residuals, and extrapolate power of ML and physical models.

## 2. Database

To establish GMPEs for predicting the PGA and PGV, data were obtained from the RotD50[34] version of the NGA-West2 flatfile[35]. Removing entries that had no listed moment magnitude ($M_w$), focal mechanism ($FM$), Joyner–Boore distance ($R_{JB}$), $V_{S30}$, or depths to the basement layer yielded a shear-wave velocity of 1.0 km/s ($Z_{1.0}$). Ground-motion records from earthquake events, for which less than five recorded excitations are available, were excluded. Earthquakes with source depths of less than 1 km and greater than 20 km were removed to ensure that the sources were located in the shallow crust. Finally, the dataset included 15175 records from 282 shallow crustal earthquakes recorded at 2655 stations with different site conditions in California, the United States, the entire Japan Island, and the Taiwan province of China (Figure 1). The recordings were selected based on $M_w$ ranging from 2.99 to 7.62, $R_{JB}$ ranging from 1.03 to 398.32 km, $V_{S30}$ ranging from 89.32 to 1464 m/s, the $FM$ (i.e., strike-slip = 1, normal-slip =2, and reverse-slip = 3), and $Z_{1.0}$ ranging from 1 to 3520 km. Figure 2 shows the distributions of PGA and PGV with distance and magnitude. A saturation phenomenon related to the magnitude and distance is indicated in both the PGA and PGV.

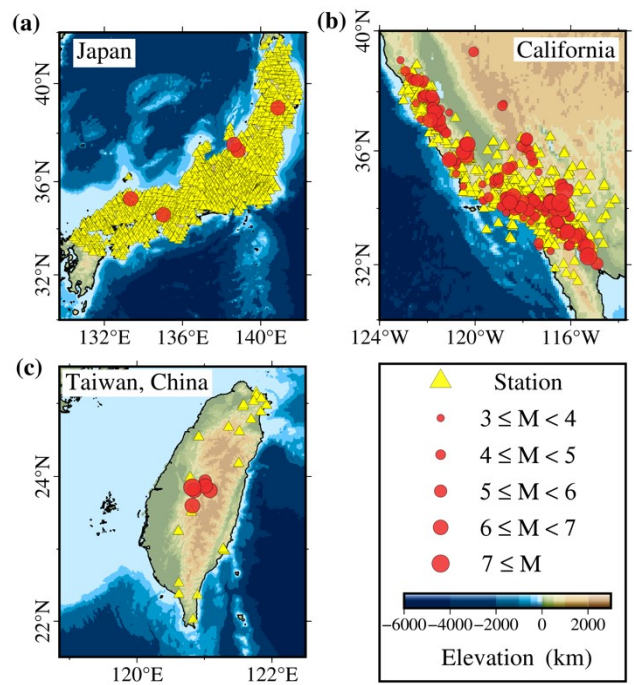

**Figure 1.** Location of strong earthquake stations and earthquake events from (a) Japan, (b) California, and (c) Taiwan, China and the magnitude distribution of these earthquake events.

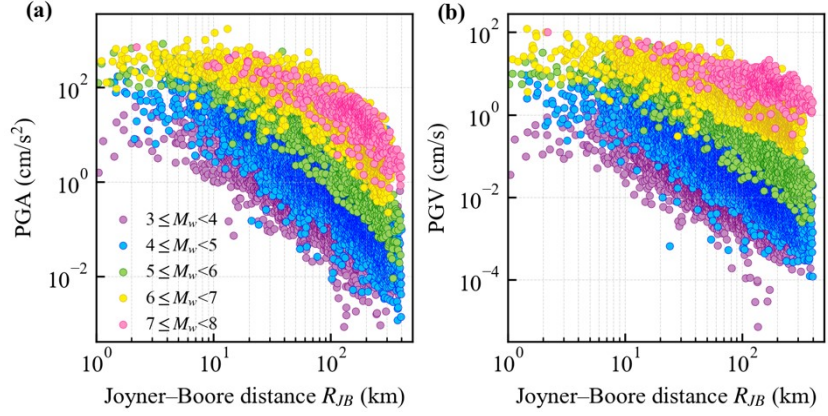

**Figure 2.** Variation in PGA (a) and PGV (b) of selected events with $R_{JB}$ in different magnitude ranges.

## 3. Proposed physics-informed symbolic learning GMMs
### 3.1 Formulation of PISL-GMMs approach

We define a sequence $\ln(Y_k) \in R^{m \times 1}$ comprising PGA ($k = 1$) or PGV ($k = 2$), where m is the number of measurements of PGA or PGV. $\Theta(X) = {1, M_w, R_{JB}, V_{S30}, \ln(M_w), \cdots, M_w^2, \cdots, FM, Z_{1.0}} \in R^{n \times m}$ is a library of *n* candidate functions containing constant, primary, logarithmic, square, and distance saturation terms, which contain parameters representing the source, path, and site. The constant term is the residual of the model and represents the degree of deviation of the model relative to the observed data. The primary, logarithmic, and squared terms represent the nonlinear attenuation of the model and are utilized in numerous GMMs[36-39]. The framework is schematically shown in Figure 3. The parameters were normalized to a uniform scale prior to model training. We assume that the constructed library of functions is competitive and contains terms related to magnitude and distance.

We embedded the distance saturation effect of the seismic near field. With reference to the elegant empirical GMPE functional form of Idriss[11,40], the distance terms associated with the magnitude, $\ln(R_{JB} + 10)$, and $M_w \ln(R_{JB} + 10)$ were added to the library (Figure 3d). This method is based on data fitting for small and moderate earthquakes with abundant data. In addition, it fully incorporates a priori physical knowledge for large earthquakes with sparse near-field data, thereby ensuring the completeness and interpretability of the function library. In addition, the effect of site amplification on the ground motion is not negligible. We referred to and simplified the description of nonlinear site amplification based on Boore[8]. We considered 1500 m/s as both the limiting velocity of $V_{S30}$ and the reference bedrock site. Therefore, we divided $V_{S30}$ by 1500 m/s before normalizing the data.

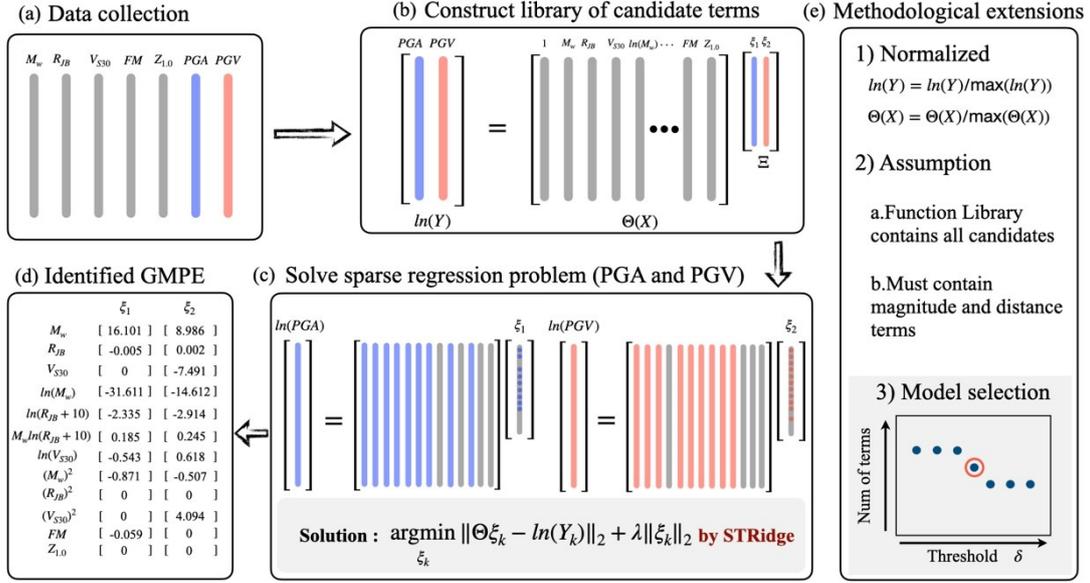

**Figure 3.** PISL-GMM framework. Arrows in the figure indicate the sequence of solutions. (a) NGA-West2 data were obtained, with each bar representing $M_w$, $R_{JB}$, $V_{S30}$, $FM$, $Z_{1.0}$, PGA, and PGV, separately; the blue and pink bars represent PGA and PGV, respectively; and all other parameters represented by gray bars. (b) A typical STRidge is constructed, where the gray columns represent the constructed candidate function library matrix; the long blue and pink columns represent ln(PGA) and ln(PGV), respectively; and the short columns represent the parameters Ξ to be solved. (c) Using the STRidge algorithm, the PGA and PGV are solved separately. (d) The calculated candidate terms and their coefficients, where $\xi_1$ and $\xi_2$ represent the coefficients corresponding to ln(PGA) and ln(PGV), respectively. (e) Some descriptions for the method, including the calculation for normalizing the data, the assumptions of the model, and a schematic diagram showing the selection of the initial thresholds.

The coefficient matrix $\xi_k \in R^{n \times 2}$ is sparse and controls the active terms in $\Theta(X)$. Hence, the following function expression is derived:

$$\ln(Y_k) = \Theta(X)\xi_k \qquad (1)$$

We assume that a sufficiently complete parameter library implies that all functional forms of the representation are fully contained in it. To obtain the optimal ξ, we transformed Equation (1) into an optimization problem with an $l^2$ norm as follows:

$$\hat{\xi} = \arg\min \| \Theta(X)\xi_k - \ln(Y_k) \|_2 + \lambda \| \xi_k \|_2, \qquad (2)$$

where $\lambda$ is the regularization factor of $\| \xi_k \|_2^2$. Rudy[30] utilized an elegant alternative method for solving Equation (2), which is known as the STRidge algorithm. The regularization factor $\lambda$ was empirically set to $10^{-7}$. The initial threshold $\delta$ determines the terms in the coefficient matrix $\xi$ that are 0. In this study, we referred to Mangan[41] to determine the regularization factor $\lambda$. The change in the number of unforced zeroing terms of the model observed with different inputs of $\delta$ is shown in Figure 4. Figure 4 shows the variation in the number of terms with $\delta$ during PGA and PGV predictions. As $\delta$ increased, the sparsity of the function library increased (i.e., the number of terms decreased). We selected the $\delta$ corresponding to the point where the number of terms decreased significantly as the optimal threshold. This guarantees both the sparsity of $\delta$ and avoids underfitting by the model.

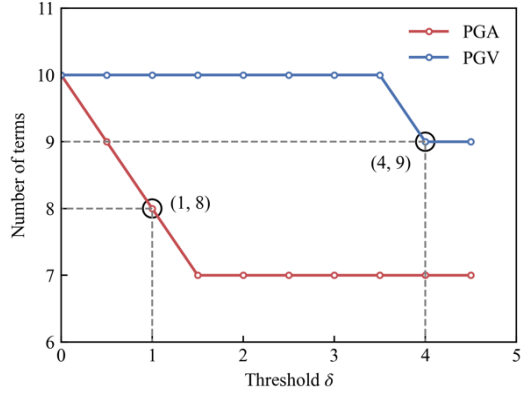

**Figure 4.** Number of terms vs. initial threshold. Red (a) and blue (b) lines indicate variation in the number of terms of the candidate function library corresponding to the training PGA and PGV models, respectively, with the threshold. White transparent circle on the line indicates the selected calculation point. Center of the black circle indicates the threshold and number of terms corresponding to the finalized ln(PGA) and ln(PGV) prediction models.

**4. Comparison of GMMs**

The PISL-GMPEs for predicting the median values of PGA and PGV are shown in Equations 3 and 4 (see Figure 3d), respectively. The PISL-GMM provide concise expressions and reflect features such as near-field saturation and anelastic attenuation. We compared PISL-GMM with leading empirical GMM[8] (BSSA14) and data-driven ANN-based GMM (ANN). The feasibility of the PISL-GMM was clarified by evaluating the residuals and extrapolation capabilities of the models.

$$\ln(PGA) = 16.101 M_w - 0.005 R_{JB} - 31.611\ln(M_w) - 0.543\ln(V_{S30}) - 0.871 M_w^2 \\ - 2.335\ln(R_{JB} + 10) + 0.185 M_w \ln(R_{JB} + 10) \quad (3)$$

$$\ln(PGV) = 8.986 M_w + 0.002 R_{JB} - 7.491 V_{S30} - 14.612\ln(M_w) + 0.618\ln(V_{S30}) - 0.507 M_w^2 + \\ 4.094 V_{S30}^2 - 2.914\ln(R_{JB} + 10) + 0.245 M_w \ln(R_{JB} + 10) \quad (4)$$

**4.1 Comparison of predicted results**

To verify the rationality of the PISL-GMPE, we compared the predicted PGA and PGV to those obtained from BSSA14 and ANN for different $M_w$, $R_{JB}$, and $V_{S30}$, as shown in Figure 5. The NGA-West2 database was used for the GMPEs.

At $V_{S30}$ = 560 and 760 m/s (Figure 5a), the PGAs predicted by all three models were similar in terms of distance and magnitude. When $V_{S30}$ = 200 m/s, the results of the three models were similar at $R_{JB} > 30$ km. However, the values predicted by the PISL-GMPE were high at $R_{JB} < 30$ km, which may be due to site nonlinearity.

At $R_{JB} < 30$ km (Figure 5b), the result yielded by the PISL-GMPE considering prior knowledge (distance saturation) was consistent with that yielded by the BSSA14; however, this was not reflected by the ANN (particularly for PGV predictions with $M_w$ = 4 and 5, see Figure 5b). Owing to the purely data-driven nature of ANNs, they may be affected by fitting bias in the absence of data or when the data distribution is uneven. At $R_{JB} > 200$ km, a clear upward bias was indicated in the predicted PGV of the PISL-GMPE, but not in that of the BSSA14 and ANN. This is similar to seismological observations. At $R_{JB} > 200$ km, the energy of the Sb phase (upper crustal diffracted S wave) was negligible, and the Sg phase (direct S wave) and Sn wave (Moho surface diffracted S wave) in the 0.5–5 s period band became the dominant energy sources of seismic S waves, which resulted in an upward shift of the PGV attenuation curve.

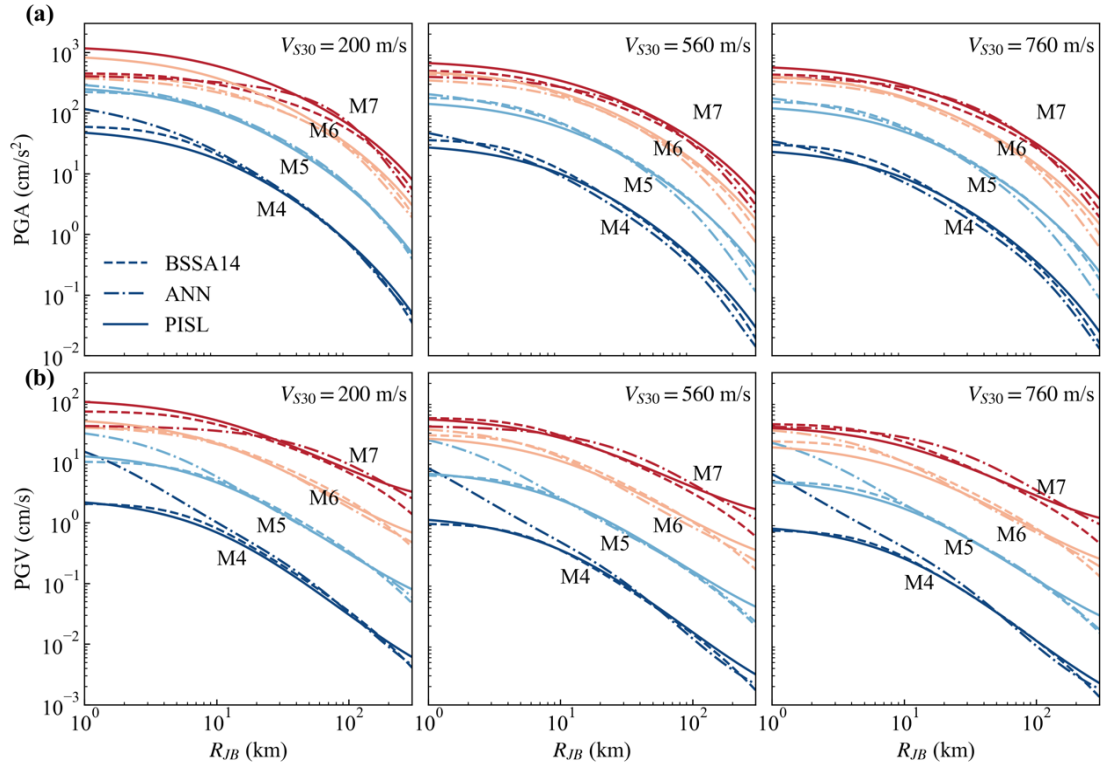

**Figure 5.** Intensity measure (PGA and PGV) to Joyner–Boore distance ($R_{JB}$) relations of GMMs determined in this study (PISL-GMM) compared with results of Boore[8] (BSSA14) and ANN. All GMM results are plotted for $V_{S30}$ = 200, 560, and 760 m/s as well as $M_w$ = 4, 5, 6, and 7.

**4.2 Residual analysis**

Residual analysis is important for assessing the dispersion degree of values predicted by the regression model. We referred to the random-effects model developed by Abrahamson and Youngs[42] for our residual analysis. The intra- and inter-event residuals of the predicted ln(PGA) and ln(PGV) vs. the measured values were calculated for the distributions of $M_w$, $R_{JB}$, and $V_{S30}$ (Figures 6 and 7). In general, the intra-event residuals of ln(PGA) were distributed between -4 and 4, and the residual values with all three variables were located near the mean value of the residuals of 0. This implies that the PGA and PGV estimated by the PISL-GMM are unbiased relative to $M_w$, $R_{JB}$, and $V_{S30}$. The residuals decreased only when $M_w$ > 7 (Figure 6f), which implies that the PISL-GMPE overestimated the observed values. This may be related to the sparsity of the large earthquake data.

The variation in the inter-event residuals with $M_w$ for the PGA and PGV is shown in Figure 7. The residuals were mainly distributed between -1.5 and 1.5, and the mean values of the residuals were located around zero. This indicates that the model demonstrated no systematic bias. Similarly, the residuals increased when $M_w$ > 7 (Figure 7), implying that the values predicted by the PISL-GMPE were not sufficiently correlated with the measured values. This may have been caused by the sparse amount of data from large earthquakes. The apparent magnitude correlation observed in the inter-event residuals may not represent the actual probability variability at large magnitudes.

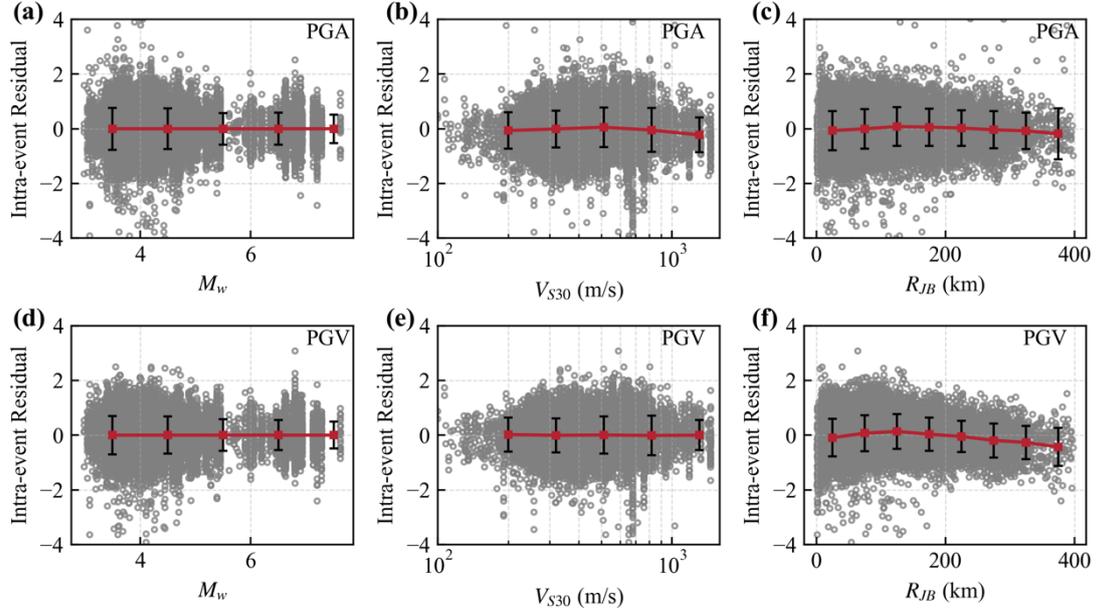

**Figure 6.** Distribution of intra-event residuals with respect to $M_w$, $R_{JB}$, and $V_{S30}$. Squares represent the mean residual and its SD in logarithmically spaced distance bins.

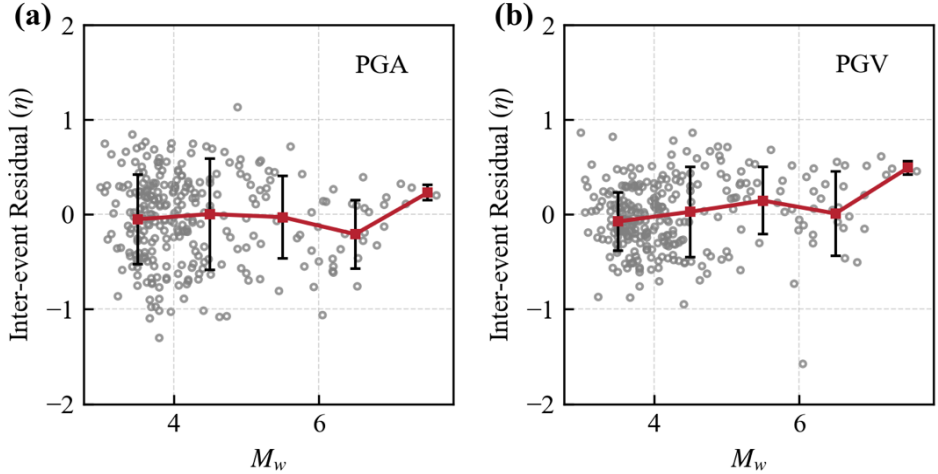

**Figure 7.** Distribution of inter-event residuals with respect to $M_w$. Squares represent the mean residual and its SD in logarithmically spaced distance bins.

To evaluate the differences among the PISL-GMPE, BBSA14, and ANN, we calculated the SD of the inter- ($\tau$) and intra-event ($\phi$) residuals for the three models based on the method of Boore[8]. To calculate $M_w$-dependent $\tau$ and $\phi$ values, the following expressions can be used:

$$\tau(M) = \begin{cases} \tau_1 & M \leq 4.5 \\ \tau_1 + (\tau_2 - \tau_1)(M - 4.5) & 4.5 < M < 5.5 \\ \tau_2 & M \geq 5.5 \end{cases} \quad (5)$$

$$\phi(M) = \begin{cases} \phi_1 & M \leq 4.5 \\ \phi_1 + (\phi_2 - \phi_1)(M - 4.5) & 4.5 < M < 5.5 \\ \phi_2 & M \geq 5.5 \end{cases} \quad (6)$$

where the values of $\tau_1$, $\tau_2$, $\phi_1$, and $\phi_2$ are listed in Table 2.

**Table 2** Values of $\tau_1$, $\tau_2$, $\phi_1$, and $\phi_2$ for the PISL-GMMs

| IM | $\tau_1$ | $\tau_2$ | $\phi_1$ | $\phi_2$ |
|---|---|---|---|---|
| PGA | 0.511 | 0.392 | 0.756 | 0.576 |
| PGV | 0.374 | 0.438 | 0.670 | 0.547 |

Using Equations 5 and 6, the inter-event residual SD $\tau$ and intra-event residual SD $\phi$ were calculated with respect to $M_w$ and compared for the three models: PISL-GMPE, BSSA14, and ANN (Figure 8). Figure 8a shows the variation in $\tau$ with the magnitude of the earthquake. Compared with the BSSA14 model, the $\tau$ of the PISL-GMPE was lower at $M_w < 4.5$ but increased at $M_w > 5.5$. Compared with the ANN model, the $\tau$ of the PISL-GMPE was lower for both $M_w < 4.5$ and $M_w > 5.5$. A comparison of the $\phi$ values in Figure 8b shows similar values for the three models, except for the increase in $\phi$ exhibited by the ANN at the PGV. The PISL-GMPE indicated a residual SD of the same order of magnitude as the other models. The model captured the variation corresponding to the magnitude, distance, and site, and is thus suitable for predicting the ground motion intensity. In addition, the three models mentioned above fitted the data well from three different perspectives: physical statistics (BSSA14), data-driven with a priori physical knowledge (PISL), and data-driven (ANN), which resulted in different variabilities among those models.

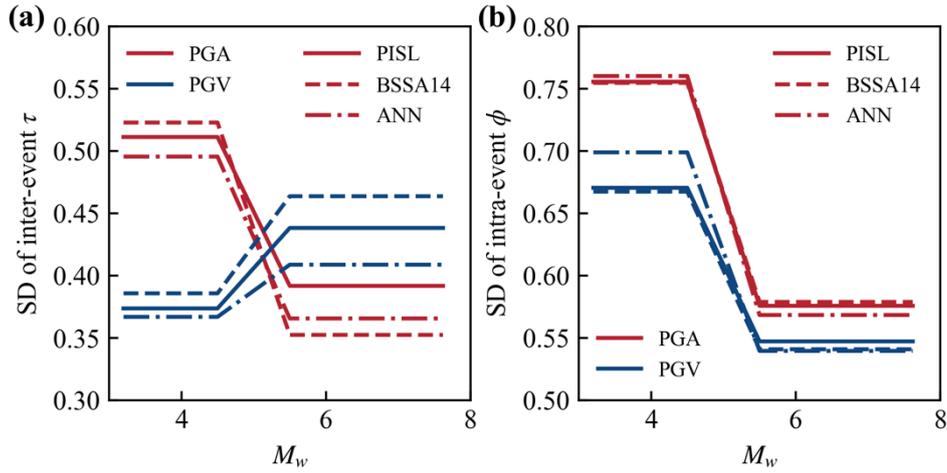

**Figure 8.** Comparison of inter-event SD $\tau$ and intra-event SD $\phi$ for three models: PISL-GMPE, BSSA14, and ANN.

### 4.3 Test of extrapolation ability

When the model was extrapolated beyond the specified range of the training data, the error became significant. Seismic hazard analysis for major infrastructures (e.g., nuclear power plants and natural gas pipelines) requires the consideration of extreme conditions. Although these cases may be outside the range of the historical earthquake data, they should not be disregarded. In particular, the data-driven GMPE is devoid of physical constraints, and its extrapolation ability must be tested. To examine the extrapolation ability of the PISL and ANN models, we retrained them using only data with $R_{JB} \geq 30$ km. Subsequently, we observed the prediction behavior of the PISL and ANN for $R_{JB} < 30$ km at different earthquake magnitudes (Figure 9). Based on the results, the PISL and ANN models indicated a similar predictive behavior at $M_w = 6$ and 7, although a clear linear trend was indicated by ANN at $M_w = 4$ and 5. The ANN model was completely data-driven. The earthquake data distribution showed a large proportion of small earthquake data (approximately 71% of the total data for $M_w < 5$) and a small proportion of large earthquake data (approximately 29% of the total data for $M_w \geq 5$). Consequently, the ANN extrapolation will result in an unrealistic data distribution. By contrast, the performance of the PISL in the near field was outstanding, which is due to the inclusion of seismic magnitude or distance saturation terms in the function library.

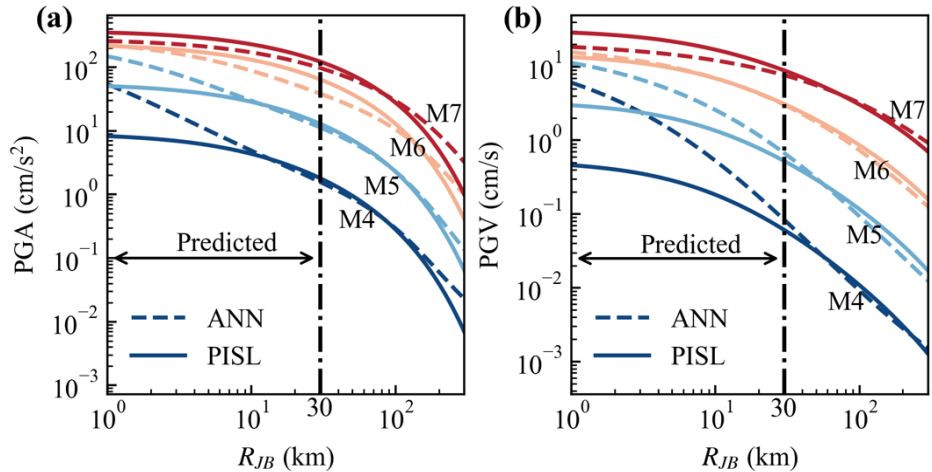

**Figure 9.** Testing the extrapolation capability of the models. The PISL and ANN models were trained using data for $R_{JB} \geqslant 30$ km to predict the performance of PGA (a) and PGV (b) for $R_{JB} < 30$ km at $V_{S30}$ = 760 m/s as well as $M_w$ = 4, 5, 6, and 7.

## 5. Conclusions

Traditional physical GMM fully integrates seismological cognition and theorems. However, implicit features in the data are typically disregarded, and the form of the assumed equations is complex. Data-driven neural-network-based models fully exploit data features but lack interpretability. The extrapolation capability of PISL-GMM and ANN was validated using only a test subset with a small amount of data. This paper outlines a physically inspired symbolic-learning GMM based on the NGA-West2 dataset. First, the stability of PISL-GMPE was demonstrated by calculating the accuracy of the models and comparing the residuals and SD of residuals with the outstanding model (BSSA14) and a deep learning model (ANN). Subsequently, the applicability of the PISL-GMPE was illustrated by comparing the variations in the PGA and PGV with $R_{JB}$ predicted using different models under different earthquake magnitudes and site conditions. Finally, the extrapolation capability of the PISL was demonstrated by comparing the prediction levels of the PISL and ANN outside the range of the training data. The results showed that the equations obtained by the PISL exhibited a concise form, possessed higher extrapolation capability than the ANN, and achieved prediction results similar to those of BSSA14.

Symbolic learning, which is an important approach in the field of knowledge discovery for artificial intelligence, has yielded favorable results in equation recognition and characterization. We applied the concept of symbols to develop GMM, which demonstrated certain advantages. Notably, in seismology and earthquake engineering, numerous measured or simulated data (e.g., soil constitutive and seismic rupture processes) can be regressed to certain display representations using symbolic learning methods. This method is beneficial to engineering applications and scientific research.